\documentclass{article}
\usepackage{spconf,amsmath,graphicx}
\usepackage{times}
\usepackage{epsfig}
\usepackage{amsmath}
\usepackage{amssymb}
\usepackage{algorithm}
\usepackage{algpseudocode}
\usepackage{varwidth}%
\usepackage{float}
\usepackage{color}
\usepackage{listings}
\usepackage{amsmath}
\usepackage{xcolor}
\usepackage{tabularx}
\usepackage{subfigure}
\usepackage{diagbox}

\title{Robust Representation Learning with Self-Distillation \\
for Domain Generalization}
\name{Ankur Singh$^{1}$, Senthilnath Jayavelu$^{1}$\thanks{This study is supported by the Accelerated Materials Development for Manufacturing Program at A*STAR via the AME Programmatic Fund by the Agency for Science, Technology and Research under Grant No. A1898b0043.}}
\address{$^{1}$Institute for Infocomm Research (I$^{2}$R), Agency for Science, Technology and Research (A*STAR)}
\begin{document}
%
\maketitle
\begin{abstract}
Despite the recent success of deep neural networks, there remains a need for effective methods to enhance domain generalization using vision transformers. In this paper, we propose a novel domain generalization technique called \textbf{R}obust \textbf{R}epresentation \textbf{L}earning with Self-\textbf{D}istillation (RRLD) comprising i) intermediate-block self-distillation and ii) augmentation guided self-distillation to improve the generalization capabilities of transformer-based models on unseen domains. This approach enables the network to learn robust and general features that are invariant to different augmentations and domain shifts while effectively mitigating overfitting to source domains. To evaluate the effectiveness of our proposed method, we perform extensive experiments on PACS \cite{li2017deeper} and OfficeHome \cite{venkateswara2017deep} benchmark datasets, as well as an industrial wafer semiconductor defect dataset \cite{wang2020deformable}. The results demonstrate that RRLD achieves robust and accurate generalization performance. We observe an average accuracy improvement in the range of \textbf{1.2\%} to \textbf{2.3\%} over the state-of-the-art on the three datasets.
\end{abstract}
\begin{keywords}
Domain Generalization, Vision Transformers,
 Self-Distillation, Data Augmentation
 \end{keywords}
\section{Introduction}
\label{sec:intro}

When training a neural network (NN), the goal is to make the model learn general patterns from the provided examples so that it can make accurate predictions on new examples it has never seen before. However, the network can sometimes learn patterns specific to the samples it was trained on and fails to generalize well to new examples. This has led to the concept of domain generalization, which aims to improve the performance of neural networks on unseen domains, but it has been a challenging problem to solve. In particular, deep neural networks (DNNs) have achieved impressive results on various benchmarks, but their performance often deteriorates when applied to new domains that are different from the training data distribution. This is a major limitation of these models, as it is often infeasible or impractical to gather large amounts of annotated data for every possible data distribution in a given application area. Retraining the model with such extensive datasets is also computationally expensive, demanding significant computational resources and time. Domain generalization, thus, becomes necessary to address these challenges.

Past research on domain generalization has mainly concentrated on enhancing the generalization performance of convolutional neural networks (CNNs). One approach is to use domain adaptation techniques, which aims to align the feature distributions of the source and target domains. A popular method for domain adaptation is the use of adversarial training \cite{Tzeng_2017_CVPR}, \cite{NEURIPS2018_ab88b157}. Another approach is to use domain-invariant feature representations, which are learned by minimizing the domain shift between different domains. A popular method for learning domain-invariant features is the use of domain-adversarial neural networks (DANN) \cite{ganin2016domain}, \cite{ghifary2015domain}. Data augmentation has also been used for domain generalization by training a model on augmented versions of the existing data, thereby making the model more robust to variations in the input data \cite{volpi2018generalizing}, \cite{zhou2020learning}. Finally, some works propose the use of a meta-learning approach to learning a model that can quickly adapt to new domains with only a few examples \cite{finn2017model}, \cite{li2018learning}. 

\begin{figure*}[t!]
\centering
\hspace{-1em}\includegraphics[width=0.82\linewidth]{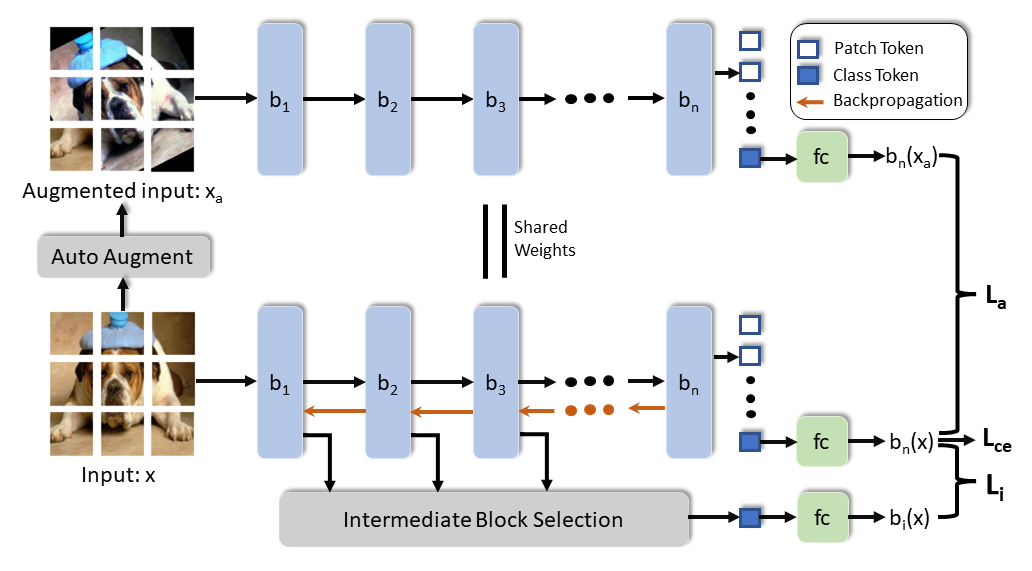}
    \caption{RRLD: The model processes an input image $x$, which is first transformed by AutoAugment to produce $x_a$. The image $x$ is then passed through the network to produce the output $b_n(x)$. A random intermediate block is selected from the network to obtain $b_i(x)$. Simultaneously, image $x_a$ is passed through the network generating $b_n(x_a)$, with gradient computation halted during this process (Refer Algorithm \ref{alg:label}). The losses are then computed between the three outputs. }
    \label{fig:model}
\end{figure*}

Recently, vision transformers (ViTs) have shown great potential in various computer vision tasks such as image classification \cite{dosovitskiy2020image}, \cite{yuan2021tokens}, object detection \cite{carion2020end}, \cite{zhu2020deformable}, semantic segmentation \cite{zheng2021rethinking}, \cite{xie2021segformer}; however, ViTs remain underexplored in the field of domain generalization. Without the ability to effectively generalize to unseen domains and data with different distributions, they may not perform optimally in real-world scenarios, thus limiting their practical deployment. 

As a result, more research is needed to develop domain generalization techniques for ViTs, addressing challenges such as domain shift, and domain-invariant feature learning.
To this extent, we propose a novel approach, Robust Representation Learning with Self-Distillation (RRLD), which addresses the aforementioned problem to improve the generalization capabilities of ViTs. The key components of our proposed method include intermediate-block self-distillation (IBSD) and augmentation-guided self-distillation (AGSD), allowing RRLD to leverage the power of data augmentation and self-knowledge distillation to enhance the generalization performance of ViTs. This technique has been proven to be effective in improving domain generalization performance, as demonstrated by our experiments on two benchmark datasets: PACS \cite{li2017deeper}, Office-Home \cite{venkateswara2017deep}, and an industrial Wafer semiconductor defect dataset \cite{wang2020deformable}. Our proposed method is an essential step towards improving the generalization capabilities of ViTs by addressing the need for effective domain generalization methods.

\section{Methodology}
\label{sec:format}

 In this section, we discuss our proposed method called 'Robust Representation Learning with Self-Distillation' (RRLD) for domain generalization. RRLD comprises: i) Intermediate-Block Self-Distillation (IBSD) and ii) Augmentation-Guided Self-Distillation (AGSD) (see Fig. \ref{fig:model}). The utilization of these components allows for improved generalization performance on unseen domains.
 


\label{sec:intro}
\subsection{Intermediate-Block Self-Distillation}
In Intermediate-Block Self-Distillation (IBSD), an intermediate block is randomly sampled from the Transformer architecture and is used to make predictions \cite{Sultana_2022_ACCV}. The image $x$ is passed through the sampled intermediate block $b_i$ where $i \in \{1, 2, \ldots, n\}$, and the classification token from the block $b_i$ is then fed into the final classifier $fc$ to produce logits $l_i$ = $b_i(x)$ as shown in Fig. \ref{fig:model}. Concurrently, we also obtain logits $l_n$ = $b_n(x)$ from the final block $b_n$ of the vision transformer network (Fig. \ref{fig:model}). Finally, we utilize KL Divergence to calculate the loss $L_{i}$ between the predictive distributions $p(l_n)$  and $p(l_{i})$ as follows:
\begin{equation} \label{Eq:Lib}
L_{i}(p(l_n)||p(l_i)) = \sum_{j=1}^{c} \sigma(l_n/T_{1})_j \log \frac{\sigma(l_n/T_{1})_j}{\sigma(l_i/T_{1})_j}
\end{equation} 
where $p(l_n)$ = $\sigma(l_n/T_{1})$, $p(l_{i})$ = $\sigma(l_i/T_{1})$,  $c$ denotes the output dimension of $fc$, $\sigma$ indicates the softmax function, and $T_{1}$ denotes the temperature scalar used to rescale the logits. 
IBSD allows us to exploit the information encoded in different parts of the network rather than relying solely on the final block. It effectively mitigates overfitting to the source domains by providing non-trivial supervisory signals for the intermediate blocks of the network, thus improving the robustness of the model. 


While IBSD brings notable benefits, it only focuses on providing supervisory signals to intermediate blocks, and further improvements are sought to ensure strong and stable learning across diverse domains. To address this, we introduce Augmentation Guided Self-Distillation, as discussed in the following subsection, to further enhance the generalization capabilities of ViTs.

\subsection{Augmentation-Guided Self-Distillation} 
The proposed self-distillation technique, Augmentation-Guided Self-Distillation (AGSD), takes advantage of the power of data augmentation and self-knowledge distillation to improve the generalization capabilities of the model. Specifically, we utilize the Imagenet policy for AutoAugment \cite{cubuk2019autoaugment} to introduce variations to the training data and then match the predictive distributions of Vision Transformers (ViTs) between the augmented and original version of the same example. This improves the generalization performance of ViTs by leveraging the knowledge gained from augmented samples of the same input.

 Notably, the input image $x$ is passed through an AutoAugment transformation, resulting in an augmented sample $x_a$. The original image $x$ processed by the network produces logits $l_n$ from the final block of the vision transformer network. The augmented sample $x_a$ is subsequently fed into the network, generating logits $l_{a_{n}}$= $b_n(x_a)$ (see Fig. \ref{fig:model}). During this process, gradient computation for the augmented samples $x_a$ is halted (refer line 7 in Algorithm \ref{alg:label}) to avoid model collapse \cite{miyato2018virtual}, \cite{yun2020regularizing}. We then use the KL divergence to compute the loss $L_{a}$ between the two predictive distributions $p(l_n)$ and $p(l_{a_{n}})$.
\begin{equation} \label{Eq: Lag}
\hspace{-0.7em}L_{a}(p(l_n) || p(l_{a_n})) = \sum_{j=1}^{c} \sigma(l_n/T_{2})_j \log \frac{\sigma(l_n/T_{2})_j}{\sigma(l_{a_n}/T_{2})_j}
\end{equation}
where $p(l_n)$ = $\sigma(l_n/T_{2})$, $p(l_{a_{n}})$ = $\sigma(l_{a_{n}}/T_{2})$, and $T_2$ denotes the temperature used to rescale the logits. 
Augmentation-guided distillation provides a way for the network to learn to generalize to a wider range of variations in the input data, which is essential for achieving good performance on different domains.  

Matching the distribution of the network between augmented versions of the same input aims to improve generalization by making the network's predictions more robust to different variations of the input. The network is trained on various augmented versions of the same example, allowing it to learn invariant representations. Furthermore, by matching the distribution of the network between augmented versions, we ensure that the network is not overfitting to specific augmentations during training, leading to better generalization on new 
data.
\begin{algorithm}[t!] \label{algorithm} 
\caption{RRLD Pytorch Pseudo Code\protect\footnotemark}\label{alg:label} 
\definecolor{codegreen}{rgb}{0.5,0.5,0.5}
\definecolor{codegray}{rgb}{0.5,0.5,0.5}
\definecolor{codepurple}{rgb}{0.58,0,0.82}
\definecolor{backcolour}{rgb}{0.95,0.95,0.92}

\lstdefinestyle{mystyle}{
  backgroundcolor=\color{backcolour}, commentstyle=\color{codegreen},
  keywordstyle=\color{magenta},
  numberstyle=\tiny\color{codegray},
  stringstyle=\color{codepurple},
  basicstyle=\ttfamily,
  breakatwhitespace=false,         
  breaklines=true,                 
  captionpos=b,                    
  keepspaces=true,                 
  numbers=left,                    
  numbersep=5pt,                  
  showspaces=false,                
  showstringspaces=false,
  showtabs=false,                  
  tabsize=2
}

\lstset{style=mystyle}
\begin{lstlisting}[language=Python, mathescape]
# N: ViT network
for x in dataloader:
    $\mathrm{x_a}$ = AutoAugment(x)
    $\mathrm{l_n}$, $\mathrm{l_{i}}$ = $\mathrm{b_n(x)}$, $\mathrm{b_{i}}$(x)
    $\mathrm{\hat{y}}$ = softmax($\mathrm{l_n}$)

    with torch.no_grad():
        $\mathrm{l_{a_n}}$ = $\mathrm{b_n(x_a)}$

    loss = $\mathrm{L_{ce}}$(y, $\mathrm{\hat{y}}$) + $\mathrm{L_{a}}$($\mathrm{p(l_n)}$, $\mathrm{p(l_{a_n})}$) 
            + 0.2*$\mathrm{L_{i}}$($\mathrm{p(l_n)}$, $\mathrm{p(l_{i})}$)
    
    loss.backward()

    update(N)

    
\end{lstlisting}
\end{algorithm}
\footnotetext{Code availability: Code for this work will be made available upon acceptance of the paper, in compliance with our organizational policies.}

\subsection{Objective function}
In our proposed method, the objective function is a combination of three loss functions: the cross-entropy loss $L_{ce}$, the intermediate-block distillation loss $L_{i}$, and the augmentation-guided distillation loss $L_{a}$.
For a given sample, the cross-entropy loss is computed between the one-hot encoded ground truth and the output obtained from the final block of the network.

The overall objective function is the sum of all three loss functions:
\begin{equation}
 L_{total} = L_{ce} + \lambda L_{i} + \gamma L_{a}
\end{equation}
where $\lambda$ and $\gamma$ are scalars multiplied to $L_{i}$ and $L_{a}$ to balance the contribution of the losses. Our proposed method allows the network to learn from multiple sources of information, such as the intermediate block and the augmented image, which improves the generalization and robustness capabilities of the network.

\section{Experiments and Results}
\label{sec:pagestyle}

\subsection{Datasets}
In the conducted experiments, we aimed to evaluate the performance of our proposed method for domain generalization on two benchmark datasets and an industrial dataset. Specifically, we conducted experiments on the PACS, and Office Home datasets, popularly used for evaluating domain generalization methods. Further, we demonstrate the effectiveness of our approach in handling out-of-distribution data by performing experiments on the Wafer dataset. 

\begin{figure}[t!]
\subfigure{\includegraphics[width=0.3\linewidth]{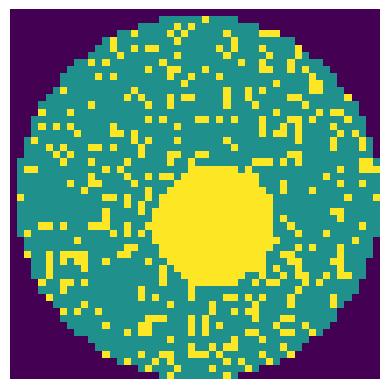}}
\hfill
\subfigure{\includegraphics[width=0.3\linewidth]{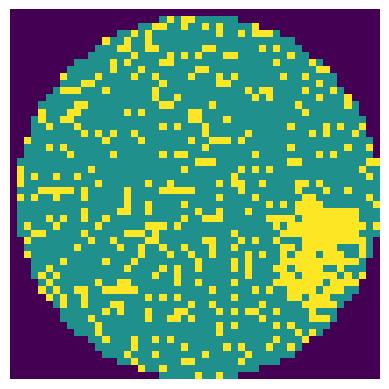}}
\hfill
\subfigure{\includegraphics[width=0.3\linewidth]{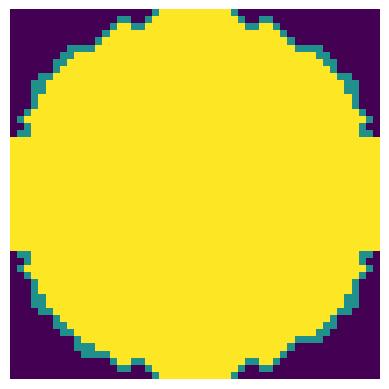}}
\caption{Wafer Dataset Images}
\label{fig:original-wafer}
\end{figure}

\begin{figure}[t!]
\subfigure{\includegraphics[width=0.3\linewidth]{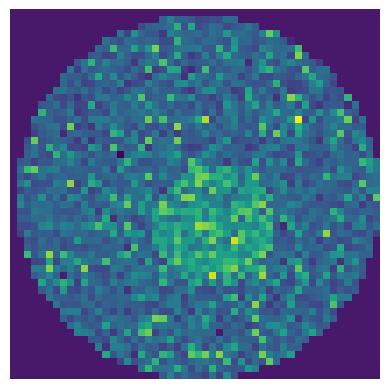}}
\hfill
\subfigure{\includegraphics[width=0.3\linewidth]{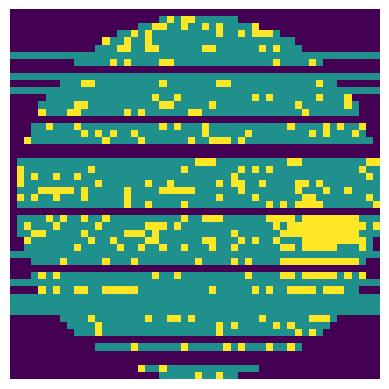}}
\hfill
\subfigure{\includegraphics[width=0.3\linewidth]{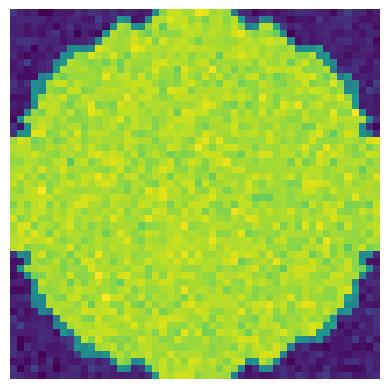}}
\caption{Noisy Wafer Dataset Images}
\label{fig:nosiy-wafer}
\end{figure}

The PACS dataset includes four domains of images: Art, Cartoons, Photos, and Sketches. It has a total of 7 classes and 9,991 images. The OfficeHome dataset also includes four domains: Art, Clipart, Product, and Real. It has a total of 65 classes and 15,588 images. To demonstrate the usefulness of our approach on an industrial dataset, we conducted experiments on a wafer semiconductor defect detection dataset. This dataset contains 8,015 images belonging to 9 classes (see Fig. \ref{fig:original-wafer}). To generate an out-of-distribution set for wafer images, we added four types of noise commonly found in semiconductor manufacturing processes (gaussian, impulse, speckle, and shot) to the images, creating a second domain (see Fig. \ref{fig:nosiy-wafer}). As a result, this dataset includes two domains - the original images and the noisy images - a total of 16,030 (8015 $\times$ 2) images. 

\subsection{Implementation} We segment the data from each training domain into training and validation sets, where 80\% is used for training and 20\% is used for validation. The validation data from each training domain are combined to create a unified validation set. The model that attains the highest accuracy on this validation set is subsequently evaluated on the target domain (test set) to report its classification accuracy. We compare the performance of two variants of Vision Transformer (ViT) models, DeiT \cite{touvron2021training}, and CvT \cite{wu2021cvt}. DeiT is a data-efficient transformer that employs knowledge distillation with a dedicated token achieving ViT-level accuracy on ImageNet using only ImageNet data. CvT integrates convolutions into the ViT architecture. We use DeiT-Small and CvT-21 in our experiments to provide a fair comparison with the work presented in \cite{Sultana_2022_ACCV}. We utilized the AdamW \cite{loshchilov2018fixing} optimizer and adopted the default hyperparameters proposed by \cite{gulrajani2020search} that include a batch size of 32 and a learning rate of 5e-05. We run our experiments three times for each test domain, using different random seeds. This allows us to obtain an average performance and also to observe any variations in the results.

In contrast to the approach used in \cite{Sultana_2022_ACCV}, where the values of $\lambda$ and $T_1$ vary for different domains and datasets, we maintain fixed values of $\lambda = 0.2$ and $T_1 = 5$. We also keep $\gamma = 1$ and $T_2 = 1$ constant throughout our experiments. By adopting fixed hyperparameters for our approach, we ensure a fair comparison of the model's performance across different datasets and domains. This uniformity eliminates any bias that might arise from varying hyperparameter settings and provides a consistent basis for evaluating the models. Furthermore, the use of fixed values for $\lambda$, $T_1$, $\gamma$, and $T_2$ facilitates a more straightforward and direct comparison, as the results are not influenced by hyperparameter tuning.

\subsection{Evaluation}
We compare our method with the state-of-the-art techniques for domain generalization utilizing vision transformer architectures as outlined in \cite{Sultana_2022_ACCV}. Specifically, we compare against the state-of-the-art method of ERM-ViT \cite{Sultana_2022_ACCV} and ERM-SDViT  \cite{Sultana_2022_ACCV}. 

\textbf{PACS}: Table \ref{table:pacs} compares the performance of the proposed method against ERM-ViT and ERM-SDViT on the PACS dataset for domain generalization. The results show that our method, RRLD, outperforms both ERM-ViT and ERM-SDViT in terms of accuracy on all domains, including Art, Cartoons, Photos, and Sketch. When using DeiT-Small as the backbone model, RRLD achieves an average accuracy of 88.8\%, outperforming ERM-SDViT by 2.5\% and ERM-ViT by a margin of 3.9\%. The results are similar when using CvT-21 as the backbone model, with RRLD achieving an average accuracy of 90.5\%, an increase of 2.2\% over ERM-SDViT and 2.7\% over ERM-ViT. These results indicate that our method, RRLD, is effective in improving the domain generalization performance on PACS.

\textbf{OfficeHome}: The OfficeHome dataset presents a bigger challenge as it consists of 65 classes. Table \ref{table:OH} compares the performance of RRLD against the other two methods. The results indicate that RRLD achieves better performance than both ERM-ViT and ERM-SDViT in terms of average accuracy across all four domains. For instance, when using DeiT-Small as the backbone model, RRLD attains an average accuracy of 73.9\%, surpassing ERM-ViT (71.4\%) and ERM-SDViT (71.5\%). Similarly, with CvT-21 as the backbone, RRLD obtains an average accuracy of 77.9\%, outperforming both ERM-ViT (75.5\%) and ERM-SDViT (75.6\%). These results affirm the efficacy of our method, RRLD, in realizing enhanced domain generalization performance on the Office-Home dataset.

\textbf{Wafer Dataset}: Table \ref{Table:2} presents the results of the proposed method, RRLD, in comparison to ERM-ViT and ERM-SDViT using the CvT-21 backbone on the Wafer dataset. It can be observed that the RRLD method outperforms both ERM-ViT and ERM-SDViT, achieving the highest accuracy of 80.6\% across three runs. ERM-ViT performed better than ERM-SDViT, attaining an accuracy of 79.4\%. These experiments demonstrate the superior performance of the proposed method across all three evaluated datasets.  

\begin{figure}[]
\center
\subfigure{\includegraphics[width=0.48\linewidth, height=3.8cm]{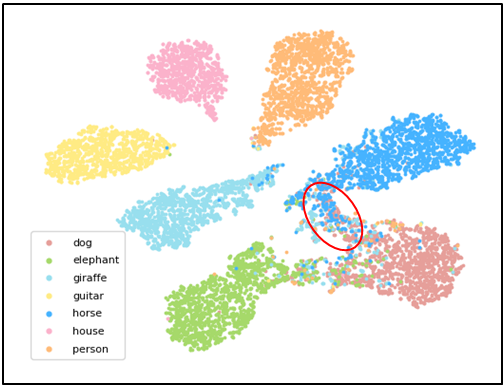}}
\hfill
\subfigure{\includegraphics[width=0.48\linewidth, height=3.8cm]{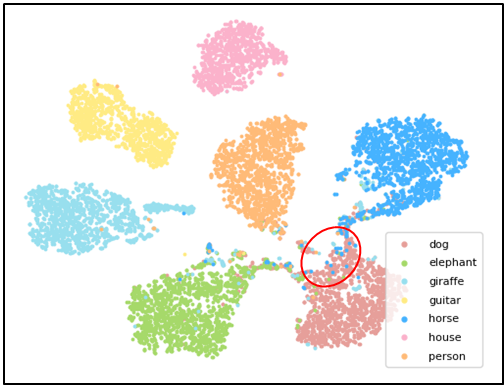}}
\hfill
 \caption{Class-wise t-SNE plots obtained from ERM-SDViT (left) and RRLD (right) on the PACS dataset. ERM-SDViT exhibits some overlap in the dog and horse classes (red regions), whereas RRLD achieves well separation between these classes (red regions).}
\label{fig:tsne_sketch_cw}
\end{figure}

\begin{figure}[]
\center
\subfigure{\includegraphics[width=0.48\linewidth, height=3.8cm]{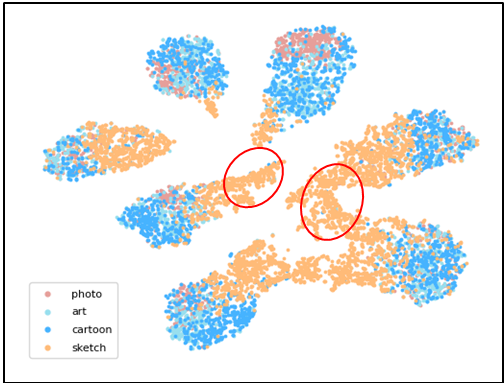}}
\hfill
\subfigure{\includegraphics[width=0.48\linewidth, height=3.8cm]{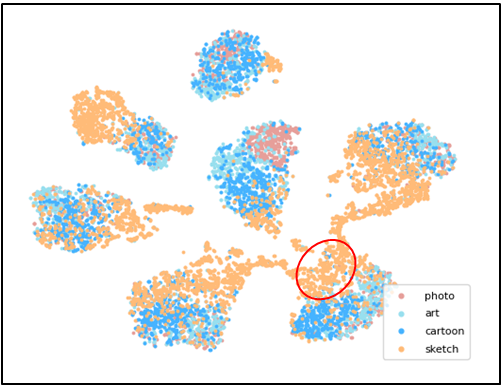}}
\hfill
 \caption{Domain-wise t-SNE plots for ERM-SDViT (left) and RRLD (right) on the PACS dataset. In the t-SNE plot of ERM-SDViT, the sketch domain is separated from the other domains (highlighted in red), indicating challenges in aligning feature representations. RRLD shows better overlap among the domains, achieving a domain-invariant feature space.}
\label{fig:tsne_sketch_dw}
\end{figure}

\begin{table*}[] 
\centering

\begin{tabular}{|l|l|l|l|l|l|l|} 
\hline
\textbf{Method}        & \textbf{Backbone}   & \textbf{Art} & \textbf{Cartoons} & \textbf{Photos} & \textbf{Sketch} & \textbf{Average} \\ \hline
ERM-ViT 
      & DeiT-Small & 87.4 $\pm$ 1.2 & 81.5 $\pm$ 0.8   & 98.1 $\pm$ 0.1 & 72.6 $\pm$ 3.3 & 84.9 $\pm$ 0.9   \\ \hline
ERM-SDViT     & DeiT-Small & 87.6  $\pm$ 0.3 & 82.4 $\pm$ 0.4 & 98.0 $\pm$ 0.3 & 77.2 $\pm$ 1.0 & 86.3 $\pm$ 0.2    \\ \hline
\textbf{RRLD} (Ours) & DeiT-Small & \textbf{90.0} $\pm$ 0.2 &   \textbf{85.0} $\pm$ 0.6 &   \textbf{98.6} $\pm$ 0.1 &   \textbf{81.7} $\pm$ 0.9    &  \textbf{88.8} $\pm$ 0.1   \\ \hline

ERM-ViT       & CvT-21     & 89.0 $\pm$ 0.1 & 84.8 $\pm$ 0.6 & 98.8 $\pm$ 0.2 & 78.6 $\pm$ 0.3 & 87.8 $\pm$ 0.1      \\ \hline
ERM-SDViT     & CvT-21     & 90.8 $\pm$  0.1 & 84.1 $\pm$ 0.5  & 98.3 $\pm$ 0.2  & 80.0 $\pm$ 1.3 & 88.3 $\pm$ 0.2  \\ \hline
\textbf{RRLD} (Ours) & CvT-21     & \textbf{91.8} $\pm$ 0.3 & \textbf{86.9} $\pm$ 0.3  & \textbf{98.9} $\pm$ 0.2  & \textbf{84.4} $\pm$ 2.0 & \textbf{90.5} $\pm$ 0.5 \\ \hline
\end{tabular}

\caption{Performance comparison on PACS.}

\label{table:pacs}
\end{table*}

\begin{table*}[] 
\centering

\begin{tabular}{|l|l|l|l|l|l|l|} 
\hline
\textbf{Method}        & \textbf{Backbone}   & \textbf{Art} & \textbf{Clipart} & \textbf{Product} & \textbf{Real World} & \textbf{Average} \\ \hline
ERM-ViT
      & DeiT-Small & 67.6 $\pm$ 0.3 & 57.0 $\pm$ 0.6   & 79.4 $\pm$ 0.1 & 81.6 $\pm$ 0.4 & 71.4 $\pm$ 0.1   \\ \hline
ERM-SDViT     & DeiT-Small & 68.3  $\pm$ 0.8 & 56.3 $\pm$ 0.2 & 79.5 $\pm$ 0.3 & 81.8 $\pm$ 0.1 & 71.5 $\pm$ 0.2    \\ \hline
\textbf{RRLD} (Ours) & DeiT-Small & \textbf{71.3} $\pm$ 0.5 &   \textbf{59.9} $\pm$ 0.7 &   \textbf{81.8} $\pm$ 0.9 &   \textbf{82.6} $\pm$ 0.2    &  \textbf{73.9} $\pm$ 0.1  \\ \hline

ERM-ViT       & CvT-21     & 74.4 $\pm$ 0.2 & 59.8 $\pm$ 0.5 & 83.5 $\pm$ 0.4 & 84.1 $\pm$ 0.2 & 75.5 $\pm$ 0.0      \\ \hline
ERM-SDViT     & CvT-21     & 73.8 $\pm$  0.6 & 60.7 $\pm$ 0.9  & 83.0 $\pm$ 0.3  & 85.0 $\pm$ 0.3 & 75.6 $\pm$ 0.2  \\ \hline

\textbf{RRLD} (Ours) & CvT-21     & \textbf{77.3} $\pm$ 0.5 & \textbf{63.2} $\pm$ 0.3  & \textbf{84.6} $\pm$ 0.7  & \textbf{86.4} $\pm$ 0.4 & \textbf{77.9} $\pm$ 0.4  \\ \hline
\end{tabular}
\caption{Performance comparison on Office Home.}

\label{table:OH}
\end{table*}

\begin{table}[]
\centering
\begin{tabular}{|m{6.5em}|m{5em}|m{6.5em}|}
\hline
\textbf{Method}        & \textbf{Backbone}   & \textbf{Wafer Dataset} \\ \hline
ERM-ViT       & CvT-21     & 79.4 $\pm$ 0.6        \\ \hline
ERM-SDViT     & CvT-21     & 78.3   $\pm$ 0.6     \\ \hline
\textbf{RRLD} (Ours) & CvT-21     & \textbf{80.6} $\pm$ 0.8      \\ \hline
\end{tabular}
\caption{Performance comparison on Wafer dataset.}
\label{Table:2}
\end{table}

\begin{table}[]
\centering
\begin{tabular}{|l|l|l|}
\hline
\textbf{Method}        & \textbf{Backbone}   & \textbf{PACS Dataset} \\ \hline
ERM-SDViT       & CvT-21 &    88.3  $\pm$ 0.2  \\ \hline
ERM-SDViT + AA    & CvT-21 &   89.5 $\pm$ 0.6     \\ \hline
    \textbf{RRLD} (Ours) & CvT-21 &     \textbf{90.5} $\pm$ 0.5        \\ \hline
\end{tabular}
\caption{Ablation Experiment to prove the effectiveness of our approach. Autoaugment improves the performance of ERM-SDViT but is unable to outperform RRLD.}
\label{table:3}
\end{table}

\begin{table*}[]
\centering
\begin{tabular}{|c|c|c|c|c|c|}
  \hline
  \diagbox[height=2em,width=5em]{\textbf{$T_{2}$}}{\textbf{$\gamma$}} & \textbf{0.1} & \textbf{0.2}& \textbf{0.5} & \textbf{1.0} & \textbf{1.5} \\ \hline
  \textbf{1} & 88.9 $\pm$ 0.6& 89.3 $\pm$ 0.7 & 89.9 $\pm$ 0.3 & \textbf{90.5} $\pm$ 0.5 & 90.1 $\pm$ 0.9 \\ \hline
  \textbf{2} & 88.5 $\pm$ 0.7 & 89.0 $\pm$ 0.5 & 89.8 $\pm$ 0.3 & 90.3 $\pm$ 0.4& 89.7 $\pm$ 0.6\\ \hline
  \textbf{5} & 88.6 $\pm$ 0.4  & 88.8 $\pm$ 0.5 & 89.1 $\pm$ 0.7 & 89.8 $\pm$ 0.5& 89.4 $\pm$ 0.4\\ \hline
  
\end{tabular}
\caption{Effect of $T_{2}$ and $\gamma$ on PACS Average Accuracy. RRLD maintains competitive accuracy levels with varying $T_2$ and $\gamma$ values. }
\label{table:t2}
\end{table*}

\subsection{t-SNE Visualization}
In Fig. \ref{fig:tsne_sketch_cw}, we present t-SNE visualizations of class-wise feature representations obtained from ERM-SDViT (left) and RRLD (right). These visualizations correspond to the  CvT-21  model trained on the photo, art, and cartoon domains, with the sketch domain serving as the target domain. Notably, the t-SNE plot for ERM-SDViT exhibits some overlap in the dog and horse classes, as indicated by the regions marked in red. In contrast, our proposed method, RRLD, showcases clear distinction and well separation between these classes, which is evident from the non-overlapping regions highlighted in red.

In Fig. \ref{fig:tsne_sketch_dw}, we visualize the domain-wise t-SNE plots for both ERM-SDViT and our proposed approach utilizing the same model as in Fig. \ref{fig:tsne_sketch_cw}. The objective of the domain-wise t-SNE plot is to observe the alignment of feature representations between different domains. Ideally, the sketch domain should overlap with other domains, signifying effective domain alignment. In the t-SNE plot of ERM-SDViT, the sketch domain is notably separated from the other domains, as indicated by the distinct cluster marked in red. This suggests that ERM-SDViT struggles to effectively align the feature representations of the sketch domain with those of other domains. Conversely, in the t-SNE plot of our proposed method, we observe a positive outcome with significant overlap among the domains. This overlap indicates that our approach successfully aligns the feature representations of different domains resulting in a domain-invariant feature space.

 \begin{figure}[t!]    \hspace{-1em}\includegraphics[width=\linewidth]{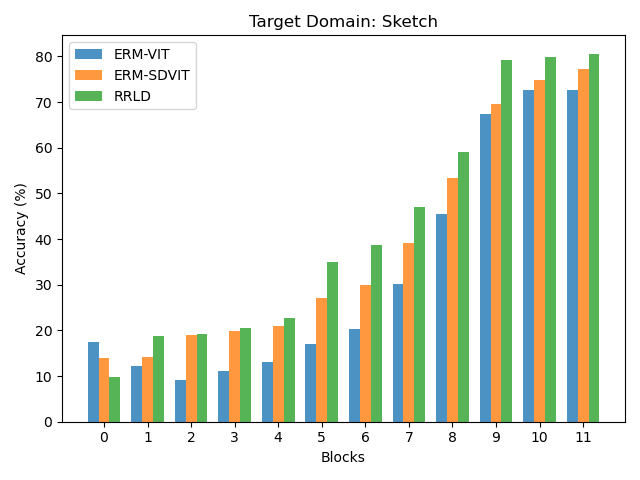}
    \caption{Block-wise accuracy of ERM-ViT \cite{gulrajani2020search}, ERM-SDViT \cite{Sultana_2022_ACCV} and our proposed method RRLD. Results are reported on the Sketch domain of the PACS dataset.}
    \label{fig:bwa}
\end{figure}




\subsection{Ablation Study}
Here we evaluate the effectiveness of augmentation-guided distillation (AGSD) in our proposed method, RRLD. We compare the performance of RRLD with ERM-SDViT and ERM-SDViT + AA, which is ERM-SDViT with the addition of AutoAugment (AA). The results are reported on the PACS dataset using the CvT-21 backbone in Table \ref{table:3}. The results of this experiment show that RRLD outperforms both methods, achieving 90.5\% accuracy, which is an improvement of 1.0\% over ERM-SDViT + AA and 2.2\% over ERM-SDViT. This demonstrates the effectiveness of augmentation-guided distillation in improving the generalization performance of our method. The results of this study also show that the use of data augmentation through the AutoAugment method can lead to improved performance of ERM-SDViT. However, it is unable to surpass the performance achieved by RRLD, which demonstrates its effectiveness in this domain generalization task. 

In Table \ref{table:t2}, we explore the impact of different temperature values $T_{2}$ and loss weightages $\gamma$ on the average accuracy of the model on the PACS dataset. For evaluation, we select the model that achieved the best validation accuracy using the specific hyperparameter setting. Notably, for $T_{2}$ = 1, $\gamma$ = 1.0, the model delivers the best average accuracy of 90.5\%. As we explore higher temperature settings, the model continues to maintain competitive accuracy levels. Similar observations can be made for different values of $\gamma$, where the model demonstrates resilience to changes in the loss weightage. For instance, at $\gamma$ = 0.1, $T_2$ = 1, the model achieves an accuracy of 88.9\%, even with a higher $\gamma$ value of 1.5, the model's accuracy remains notable at 90.1\%.

\subsection{Intermediate Block Performance}
In Fig. \ref{fig:bwa}, we assess ERM-ViT, ERM-SDViT, and RRLD for their intermediate block accuracies. For a fair comparison, the experiments were performed using the DeiT-Small model on the sketch domain of the PACS dataset, with the settings being the same as those used in \cite{Sultana_2022_ACCV}. ERM-SDViT notably enhances accuracy across intermediate blocks over ERM-ViT. However, RRLD, incorporating intermediate block self-distillation and augmentation-guided self-distillation, outperforms ERM-SDViT, demonstrating superior accuracy enhancement. 

\section{Conclusion}
In this paper, we proposed a novel domain generalization approach called RRLD. Our proposed method consists of augmentation-guided self-distillation and intermediate block self-distillation, which uses a transformer network to learn domain-invariant features. We evaluated RRLD on three domain generalization datasets and compared its performance to state-of-the-art methods. Our results show that RRLD outperforms existing methods, yielding notable improvements of 2.1\% on PACS, 2.3\% on Office Home, and 1.2\% on the Wafer dataset compared to the state-of-the-art methods. Our extensive experiments demonstrate the effectiveness of RRLD in achieving robust and accurate generalization performance. The proposed method is simple yet effective, and it could be helpful for a wide range of domain generalization tasks.

\pagebreak

{\fontsize{8.5}{10}\selectfont
\bibliographystyle{IEEEbib}
\bibliography{refs, strings}
}

\end{document}